\pdfoutput=1

\documentclass[11pt]{article}

\usepackage[final]{acl}
\usepackage{multirow}
\usepackage{comment}

\usepackage{times}
\usepackage{latexsym}
\usepackage{amsmath,bm}
\usepackage{booktabs}
\usepackage{amsfonts}
\usepackage{tabularx}

\usepackage[T1]{fontenc}

\usepackage[utf8]{inputenc}

\usepackage{microtype}

\usepackage{inconsolata}

\usepackage{graphicx}

\title{Cross-Lingual Transfer Learning for Speech Translation}

\author{Rao Ma, Mengjie Qian, Yassir Fathullah, Siyuan Tang, Mark Gales, Kate Knill \\
  ALTA Institute, Department of Engineering, University of Cambridge \\
  \texttt{\{rm2114,mq227,yf286,st941\}@cam.ac.uk, mjfg@eng.cam.ac.uk, kmk1001@cam.ac.uk} \\}

\begin{document}
\maketitle
\begin{abstract}
There has been increasing interest in building multilingual foundation models for NLP and speech research. This paper examines how to expand the speech translation capability of these models with restricted data. Whisper, a speech foundation model with strong performance on speech recognition and English translation, is used as the example model. Using speech-to-speech retrieval to analyse the audio representations generated by the encoder, we show that utterances from different languages are mapped to a shared semantic space. This shared embedding space can then be leveraged for zero-shot cross-lingual transfer in speech translation. By fine-tuning the Whisper decoder with only English-to-Chinese speech translation data, improved performance for translation to Chinese can be obtained for multiple languages, in addition to English. Furthermore, for languages related to those seen in training it is possible to perform speech translation, despite the model never seeing the language in training, or being able to perform transcription.



\end{abstract}

\section{Introduction}
Speech translation (ST) systems directly generate transcriptions in the target language from spoken utterances in a different language and have various applications~\cite{inaguma2019multilingual, nakamura2009overcoming}. 
With the growing demand for multilingual models, it is crucial to develop translation systems that support multiple languages, both as source and target. 
However, data collection for training ST systems is more challenging than for Neural Machine Translation (NMT) and Automatic Speech Recognition (ASR) tasks. Unlike NMT, where the same text corpus can be used for both translation directions~\cite{artetxe2019massively}, ST systems face challenges due to their asymmetric input-output nature. For instance, data for translating audio in language $X$ into text in English ($X\!\rightarrow\!en$) would be easier to collect than $en\!\rightarrow\!X$ data, largely due to the higher global demand for English translations. Moreover, high-resource language pairs have more available data than low-resource pairs.

Given the high cost of collecting diverse data pairs for ST systems, understanding what is required to build a multilingual ST model and expand its capability to more languages is essential. In this work, we use OpenAI's Whisper~\cite{radford2023robust} as a case study to explore the behavior of multilingual speech foundation models. 
Whisper is pre-trained to support speech recognition in 100 languages and translation from 99 languages into English ($X\!\rightarrow\!en$). 
The encoder can extract semantic information from the acoustic features. We hypothesise that the features in different languages are aligned within a shared semantic space, and this alignment could enable the model to support translation from multiple source languages, a key feature for expanding multilingual ST capabilities.
Whisper's decoder acts as a language model that generates tokens conditioned on the encoder outputs. By supporting multiple languages at the token level, the decoder facilitates translation into various target languages. This flexibility allows us to test and expand its ST capabilities to new target languages, which we verify through zero-shot and fine-tuning experiments. 


In this work, we explore how to extend Whisper's capability in speech translation,  expanding its supported translation language pairs. First, we evaluate the level of language invariance in the embeddings produced by the Whisper encoder using a speech-to-speech retrieval task~\cite{lee2015spoken}. Second, we expand the translation to a new target language by fine-tuning Whisper, the results show a level of cross-lingual transferability among the source languages. Third, we show that Whisper can translate spoken utterances from previously unseen languages into English texts, indicating its ability to map unseen languages into a shared speech embedding space.




\begin{figure*}[!htbp]
    \centering
    \includegraphics[width=0.98\linewidth]{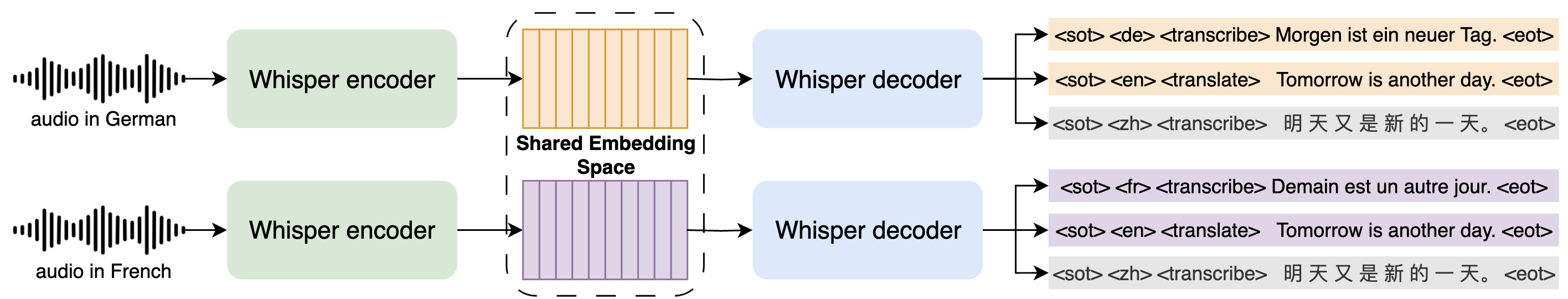}
    \caption{Illustration of Whisper's decoding process for ASR and speech translation tasks. Whisper supports speech recognition in 100 languages and speech translation from any language into English (orange (German, $de$, input) and purple (French, $fr$, input) text blocks). Fine-tuning on English-to-Chinese, $en\!\rightarrow\!zh$, speech translation data enables the model to acquire additional speech translation capabilities (such as $de\!\rightarrow\!zh$ and $fr\!\rightarrow\!zh$) through cross-lingual transfer (gray text blocks). The Whisper {\texttt{\textless transcribe\textgreater}} task token is used in this case as the {\texttt{\textless translate\textgreater}} task token causes English words to be output, independent of the target language.}
    \label{fig:whisper}
\end{figure*}


\section{Related Works}
\label{sec:related}


Prior work has shown that multilingual text models, such as M-BERT \cite{pires2019multilingual}, produce language-invariant embeddings, mapping the same semantic information from different languages to a similar embedding space. This language invariance enables cross-lingual text retrieval~\cite{pires2019multilingual, wu2019beto, cao2020multilingual} and boosts the model performance in other languages, when fine-tuned only on English corpus~\cite{pires2019multilingual}. This transfer learning capability is particularly beneficial in low-resource settings. \cite{schwenk2017learning, artetxe2019massively} have shown that using machine translation as the training objective can effectively generate language-invariant embeddings.


Unlike text models, Whisper's pre-training for speech translation only uses English as the target language.
Recently, \cite{peng23d_interspeech} have demonstrated that Whisper exhibits emergent capabilities in unseen speech translation directions through prompt engineering at inference for  ($en\!\rightarrow\!X$) speech translation.
In this study, we conduct a more comprehensive investigation into Whisper's cross-lingual transferability.

Whisper's utterance embeddings are not explicitly aggregated, again unlike text models. Additionally, speech representations are much longer than text tokens. These differences add to the difficulty of auto-alignment in the speech encoder space.
In the speech area, \cite{khurana2022samu} learned multimodal multilingual speech embeddings by fine-tuning from a pre-trained XLS-R model \cite{babu2022xls}. They used the LaBSE text encoder model \cite{feng2022language}, which produces aligned embedding spaces across languages, as the teacher model during training. For each given language, the proposed SAMU-XLSR model generates utterance-level speech embeddings and is trained to minimize the cosine loss relative to the teacher model's output. Through knowledge distillation, the model can produce an aligned speech embedding space. \cite{duquenne2021multimodal, duquenne2023sonar} followed a similar idea to align the space produced by the speech encoder with a pre-trained multilingual text encoder. Our work differs in that Whisper is not explicitly trained to match a text encoder space; instead, we rely on the speech translation pre-training target to achieve automatic alignment. Moreover, Whisper generates speech embeddings at a frame-level granularity rather than at the utterance level, enabling more fine-grained representations.


\section{Speech Translation}
\label{sec:st}
\subsection{Whisper Model}
The Whisper models are trained in a weakly supervised way and come in various sizes, from the tiny model with 39M parameters to the large model with 1550M parameters \cite{radford2023robust}. 
During pre-training, the model learns in a multi-task fashion on automatic speech recognition, speech translation, voice activity detection, and language identification. 
In decoding, it generates different outputs based on the ``context'' tokens given to the decoder. For ASR, Whisper converts an utterance in language \texttt{L} into its corresponding transcription, $\text{Utt}_\texttt{L}\!\rightarrow\!\text{Text}_\texttt{L}$. For speech translation, it supports translation from any supported language to English, represented as $\text{Utt}_\texttt{L}\!\rightarrow\!\text{Text}_\texttt{EN}$. Figure \ref{fig:whisper} shows an example of the standard transcription and translation decoding processes and the associated context tokens in orange and purple text blocks.


\subsection{Audio Embeddings}
\label{sec:embedding}
Given that multilingual text models like M-BERT generate language-invariant embeddings, it is reasonable to investigate whether Whisper, a multilingual speech model, exhibits similar properties. If Whisper's encoder produces language-invariant speech embeddings, it would be a significant advantage for handling multiple source languages in speech translation. This cross-lingual capability enables Whisper to effectively translate between various language pairs by aligning speech representations across different source languages.

To assess the cross-lingual alignment of Whisper, we use zero-shot speech-to-speech retrieval tasks~\cite{boito2020mass, duquenne2023sonar} as an evaluation method. 
In this task, given a query audio $q$, the goal is to retrieve an utterance $\hat{r}_q$ in the target language that conveys the same meaning as $q$ from a set of $R$ candidates. 
We measure the performance of the speech retrieval task using the recall rate, 
$\text{R@1}=\frac{1}{|Q|}\sum_{q\in Q} \mathbb{I}(r_q, \hat{r}_q)$
where $r_q$ is the retrieved result and $\hat{r}_q$ is the reference.
For each query $q$ and candidate audio $r$, we extract the encoder output sequences from Whisper, denoted as $E_q$ and $E_r$. The retrieved utterance $r_q$ is then determined as the one with the highest similarity score,
$r_q = \mathop{\arg\max}_{r\in R} \text{Sim}(E_q, E_r)$.

We propose \textbf{SeqSim}, a metric inspired by BERTScore~\cite{zhang2019bertscore}, to compute similarity  between two speech embedding sequences:
\begin{eqnarray}
\small
\begin{split}
    &\text{Re}_{\mathrm{seq}}=\frac{1}{|X|} \hspace{-0.02in} \sum_{\bm x \in X} \hspace{-0.02in} \max _{\bm y \in Y} {\bm{x}}^{\hspace{-0.02in}\top} \hspace{-0.03in}\bm y; 
    \hspace{-0.1in}
    \quad \text{Pr}_{\mathrm{seq}}=\frac{1}{|Y|} \hspace{-0.02in} \sum_{\bm y \in Y} \hspace{-0.02in} \max _{\bm x \in X} \bm{x}^{\hspace{-0.02in}\top} \hspace{-0.03in}{\bm{y}} 
    \\
    & \hspace{0.8in}
    \text{SeqSim}=2 \cdot \frac{\text{Pr}_{\mathrm{seq}} \cdot \text{Re}_{\mathrm{seq}}}{\text{Pr}_{\mathrm{seq}}+\text{Re}_{\mathrm{seq}}}
\end{split}
\end{eqnarray}
While BERTScore evaluates text generation tasks by comparing embeddings of individual tokens, SeqSim adapts this concept for audio frames. It computes the cosine similarity between embeddings of audio frames from one speech utterance $\bm X$ and those from another speech utterance $\bm Y$. Specifically, SeqSim measures how well each audio frame in $\bm X$ matches with the most similar frame in $\bm Y$.

\subsection{New Target Languages}
Although Whisper was trained to translate speech into English, its decoder has been exposed to a diverse range of languages and their corresponding tokens throughout its training for the transcription task. This extensive multilingual exposure suggests that the model might also be capable of translating into other languages.
To investigate this potential, we evaluate Whisper's baseline translation performance for languages beyond English. Following~\cite{peng23d_interspeech}, which demonstrated that the {\texttt{\textless transcribe\textgreater}} task token can outperform {\texttt{\textless translate\textgreater}} in the translation task, we compare these tokens in the zero-shot experiments to test translation into new target languages. Fine-tuning the model for a new target language is also compared.
Figure~\ref{fig:whisper} shows the decoding process with an added target language: Chinese, $zh$.

Whisper's pre-training on multilingual speech enables it to generate embeddings in a shared semantic space, promoting cross-lingual transferability.
This feature allows Whisper to handle multiple source languages in speech translation. When fine-tuning Whisper for a specific language pair to expand the speech translation to a new target language (e.g. $en\!\rightarrow\!zh$), we expect improved performance for other source languages translating into the same target language ($X\!\rightarrow\!zh$). This aspect will be examined in Section~\ref{sec:exp_st_target}.

\subsection{New Source Language}
Low-resource languages not seen during Whisper’s training have different lexical representations compared to the languages the model was trained on. However, they may share similar acoustic features.
It remains to be seen whether speech embeddings for these low-resource languages also fall within the model's shared semantic space. If so, this alignment could enable Whisper to effectively expand its speech translation capabilities to include these new source languages. Section~\ref{sec:sec:exp_st_source} will explore this possibility through experiments.



%

\section{Experimental Results}
\label{sec:expres}
\subsection{Setup}
The Whisper large-v2 model is selected for the multilingual speech translation experiments, which shows superior performance compared to other model sizes \cite{radford2023robust}. We evaluate speech translation on the FLEURS dataset \cite{conneau2023fleurs}, which provides n-way parallel speech data. For the main experiments, we selected 5 languages: English (en), French (fr), German (de), Chinese (zh), and Japanese (ja). These were chosen for their wide usage and representation of different language families. To extend Whisper's ability to translate into a new target language, we use the en-to-zh subset from the CoVoST 2 dataset \cite{wang2021covost}, totalling 428 hours, in supervised training. For experiments in Section \ref{sec:sec:exp_st_source} evaluating new source languages, we choose 6 languages unsupported by Whisper: Kabuverdianu (kea), Asturian (ast), Cebuano (ceb), Kyrgyz (ky), Sorani Kurdish (ckb), and Irish (ga). Detailed descriptions of the datasets and the experimental setup are provided in Appendix \ref{app:data} and \ref{app:train}.





\subsection{Results on Speech-to-Speech Retrieval}
\label{sec:exp_retrieval}
\begin{table}[!t]
\small
    \centering
    \begin{tabular}{c|ccccc}
    \toprule
        \multirow{2}*{Query} & \multicolumn{5}{c}{R@1 [\%]} \\
        & en & fr & de & zh & ja \\
        \midrule
        en & - & 80.0 & 80.0 & 46.2 & 45.5  \\
        fr & 73.2 & - & 64.8 & 42.0 & 48.1 \\
        de & 70.4 & 62.2 & - & 42.7 & 48.1 \\
        zh & 26.5 & 25.4 & 19.0 & - & 43.2 \\
        ja & 18.1 & 22.3 & 16.4 & 35.2 & - \\
    \bottomrule
    \end{tabular}
    \caption{Zero-shot speech-to-speech retrieval results measured with SeqSim on FLEURS.}
    \label{tab:fleurs}
\end{table}
\noindent
In preliminary experiments, we compared various similarity measures on three language pairs from FLEURS. SeqSim consistently outperformed other measures in capturing speech embedding similarity. Consequently, SeqSim is adopted for the retrieval experiments presented in this paper. A detailed comparison and results are discussed in Appendix~\ref{app:comp_sim_measure}.

Using SeqSim,  we conduct experiments on 20 language pairs from the FLEURS dataset, with results detailed in Table~\ref{tab:fleurs}. On all 20 language pairs, SeqSim consistently achieved remarkably higher recall rates compared to a random baseline of 0.2\%. This suggests that these languages share a common embedding space, where semantically similar speech utterances are mapped to close regions.
Notably, retrieval performance is better when both the query and the candidate utterances belong to the same language family. For instance, retrieval between English (en), French (fr), and German (de) -- all Indo-European languages -- shows higher performance. This is likely due to greater overlap in phoneme representations among these languages, which facilitates the model's ability to align and match audio frames effectively.

\subsection{New Target Language}
\label{sec:exp_st_target}

Whisper is originally designed for speech translation into English. This section explores methods to extend its capabilities to translate into other target languages, using Chinese as an example.

\subsubsection{Zero-shot}
As demonstrated in \cite{peng23d_interspeech}, modifying the default special tokens provided to the decoder enhances Whisper's zero-shot speech translation performance on unseen languages. Following this work, we tested two sets of context tokens in the zero-shot experiments: {\texttt{\textless sot\textgreater\textless zh\textgreater\textless translate\textgreater}} and \texttt{\textless sot\textgreater\textless zh\textgreater\textless transcribe\textgreater}. The first set follows Whisper's default speech translation decoding process. Since Whisper was initially trained to produce English translations, it outputs English words even when the target language code $zh$ is used. In contrast, utilizing the \texttt{transcribe} token resulted in a significant performance improvement, as shown in Table~\ref{tab:st}, with performance gains comparable to those reported in \cite{peng23d_interspeech}. This suggests that Whisper has learned to handle tokens of multiple languages through its multilingual speech recognition training, suggesting its potential for translating into languages beyond English.
\subsubsection{Fine-tune}
We fine-tune Whisper on English-to-Chinese speech translation data from CoVoST, freezing the encoder to preserve the audio embedding space and updating only decoder parameters with the context tokens \texttt{\textless sot\textgreater\textless zh\textgreater\textless transcribe\textgreater}. This improved English-to-Chinese translation on the FLEURS and CoVoST 2 datasets, as shown in Table~\ref{tab:st}. Testing French, German and Japanese utterances from FLEURS revealed that fine-tuning also improved BLEU and COMET scores for these languages. Although these source languages were not included in fine-tuning, the improvements in English translation capabilities benefited them due to the cross-lingual alignment feature of Whisper.

\begin{table}[!t]
    \centering
    \small
    \begin{tabular}{@{ }cc|cc|c@{ }}
    \toprule
    \multicolumn{2}{c|}{BLEU / COMET} & \multicolumn{2}{c|}{Zero-shot} & Fine-tune  \\ \cline{1-2}
    Dataset & src &  Translate & Transcribe & en-to-zh \\
    \midrule
    \multirow{4}*{FLEURS} & en & 1.0 / 58.8 & 10.3 / 66.3 & \textbf{29.1} / \textbf{78.4} \\
        & fr & 0.9 / 56.2 & 15.7 / 66.7 & \textbf{23.0 }/ \textbf{74.1} \\
        & de & 1.0 / 57.2 & 16.8 / 67.1 & \textbf{24.0} / \textbf{74.7} \\
        & ja & 1.0 / 59.3 & 15.9 / 70.7 & \textbf{19.2} / \textbf{74.7} \\
    \midrule
        CoVoST 2 & en & 1.8 / 59.0 & 3.8 / 61.2 & \textbf{31.9} / \textbf{76.3} \\
    \bottomrule
    \end{tabular}
    \caption{Zero-shot and fine-tuning results (BLEU / COMET) for Whisper speech translation into Chinese.}
    \label{tab:st}
\end{table}

\subsection{New Source Languages}
\label{sec:sec:exp_st_source}

We have shown that Whisper features a shared semantic embedding space across languages. This section explores whether this cross-lingual transferability extends to low-resource languages that Whisper has not been directly trained on. To test this, we select 6 unsupported languages from the FLEURS dataset and used a language code from their most similar language (chosen based on vocabulary overlap) for decoding~\cite{qian2024learn}. 
Whilst Whisper struggles with accurate ASR transcriptions for these low-resource languages, as shown by the high WER in Table~\ref{tab:low}, some languages exhibit high recall (R@1) rates when retrieving English speech (such as Kabuverdianu (kea) and Asturian (ast)).  This suggests that even though these languages were unseen during training, their audio embeddings are mapped to the shared semantic space. This effectiveness likely results from the audio similarities between these low-resource languages and those in Whisper's training data.

Utilising these speech embeddings, the Whisper decoder can translate these languages into English.  The results in Table~\ref{tab:low} reveal surprisingly good BLEU scores for languages like Kabuverdianu and Asturian (only BLEU scores are given as some languages are not supported by COMET). This suggests that Whisper’s cross-lingual alignment enhances performance in both retrieval and translation tasks for languages not explicitly included in its training.



\begin{table}[!t]
\small
    \centering
    \begin{tabular}{lc|c|c|c}
    \toprule
        src & code & WER & R@1 & ST (en) \\
    \midrule
        kea & pt & 89.5 & 85.4 & 32.6 \\
        ast & es & 47.8 & 72.8 & 27.9 \\
        ceb & en & 98.1 & 37.9 & 10.0  \\
        ky & ru & 103.2 & 21.0 & 4.2  \\
        ckb & fa & 107.1 & 19.1 & 1.9  \\
        ga & en & 105.9 & 11.0 & 2.6  \\
    \bottomrule
    \end{tabular}
    \caption{ASR, retrieval (R@1), and ST (BLEU score) into English for 6 unsupported languages on FLEURS data, with Whisper decoding language code specified.}
    \label{tab:low}
\end{table}

\section{Conclusions}
This work demonstrates how to extend speech translation capabilities in Whisper. Whisper’s decoder, supporting diverse language tokens, allows for effective expansion to new target languages. Our experiments reveal high recall rates in speech-to-speech retrieval, indicating that Whisper’s encoder captures language-invariant features across languages. Fine-tuning Whisper on English-to-Chinese ($en \rightarrow zh$) data improved BLEU scores by 5.9 for three other source languages. In addition, Whisper can successfully translate speech from some previously unseen languages into English, despite high WERs. These results confirm that Whisper maps utterances into a shared embedding space, enabling effective cross-lingual transfer for speech translation.



\section*{Acknowledgments}

This paper reports on research supported by Cambridge University Press \& Assessment, a department of The Chancellor, Masters, and Scholars of the University of Cambridge. Mengjie Qian was partially supported by EPSRC Project EP/V006223/1 (Multimodal Video Search by Examples).

\section{Limitations}
Despite promising results, this work has several limitations. First, fine-tuning Whisper on $en \rightarrow zh$ translation data led to performance degradation on $X\rightarrow en$ translations, highlighting a common issue of catastrophic forgetting. Additionally, our experiments mainly focused on one new target language. While we believe the findings are applicable to other target languages, evaluating the model across a broader range of target languages would provide a more comprehensive assessment of its capabilities. Lastly, although Whisper shows potential for unseen languages, there is room for improvement in handling low-resource languages more effectively, such as Irish (ga). Future work will explore these aspects.

\section{Risks and Ethics}
There are no known ethical concerns or risks associated with the findings of this work.



\bibliography{custom}

\clearpage
\appendix
\section{Experimental Setup}
\label{sec:appendix}

\subsection{Data Details}
\label{app:data}

Table \ref{tab:data} listed three public datasets we used in the experiments. For the FLEURS dataset \cite{conneau2023fleurs}, we processed the data by retaining only the utterances that are available in all five selected languages. The original dev and test sets provided in the dataset are combined to create a bigger evaluation set. To increase the difficulty of the designed retrieval task, we randomly kept only one instance for utterances with the same transcription but recorded by different speakers. For the supervised experiments, we fine-tune the Whisper model on the CoVoST 2 dataset \cite{wang2021covost}, which is part of the Common Voice project \cite{ardila2020common}. In the speech retrieval experiments to demonstrate the alignment of the encoder outputs, an additional dataset MaSS \cite{boito2020mass} is used. The MaSS dataset contains parallel speech data extracted from verses in 8 languages: English (en), Spanish (es), Russian (ru), Romanian (ro), French (fr), Finnish (fi), Hungarian (hu), and Basque (eu). As the released Hungarian data is incomplete we discarded it in the experiments. 

\begin{table}[!htbp]
\center
\small
\begin{tabular}{ll|cccc}
\toprule 
Dataset & Split & Langs & Utts & Hours & Words \\
\midrule
FLEURS & test & 5 & 426 & 1.1 & 9K \\
\midrule
\multirow{3}*{CoVoST} & train & 2 & 288,204 & 428 & 2.8M \\
& dev &2 & 1,000 & 1.6 & 9K \\
& test &2 & 1,000 & 1.6 & 9K \\
\midrule
MaSS & test & 7 & 814 & 8.3 & 18K \\
\bottomrule
\end{tabular}
\caption{Dataset description. The number of utterances, total duration of speech data, and word counts in the references are calculated based on the English data.} 
\label{tab:data}
\end{table}

\subsection{Training Details}
\label{app:train}
In the training and evaluation of Whisper, the original audio is chunked or padded into segments with a length of 30 seconds. In our zero-shot speech-to-speech retrieval experiments, we only keep the embedding vectors that correspond to meaningful content in the original audio and remove the ones associated with the padded part. This practice proves to be effective in the retrieval experiments. To evaluate the model performance on ST, we use BLEU \cite{papineni2002bleu} and COMET scores \cite{rei2020comet,stewart-etal-2020-comet,rei-etal-2022-comet} with the \textit{Unbabel/wmt22-comet-da} model. In the supervised ST setting, the model parameters are updated on the training set of CoVoST 2 for 220K steps with fine-tuning or LoRA tuning \cite{hu2021lora}. The initial learning rate is $1e^{-5}$ for fine-tuning and $1e^{-3}$ for LoRA tuning and decays linearly. A batch size of 16 is used during training. 

\section{Analysis of Audio Embeddings}

\subsection{Visualisation of Encoder Alignment}

\begin{figure}[!htbp]
    \centering
    \includegraphics[width=0.89\linewidth]{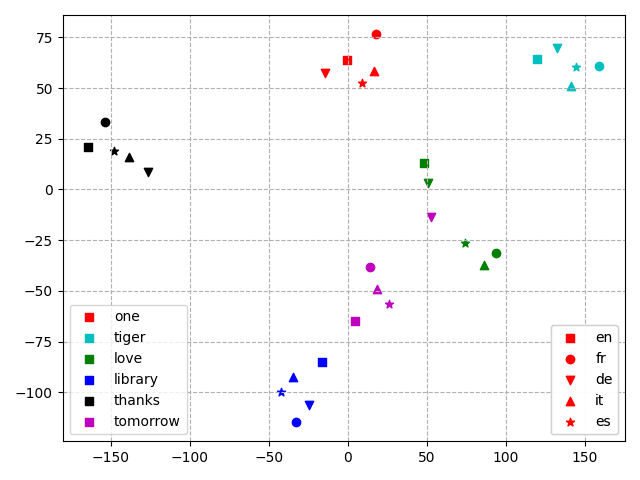}
    \caption{t-SNE visualization of contextual speech embeddings generated by Whisper large-v2 encoder for 6 word tuples across 5 languages.}
    \label{fig:tsne}
\end{figure}

\noindent
To study the language-invariance of the Whisper encoder space, we use the Amazon text-to-speech service~\cite{lorenzotrueba2019towards,klimkov2019fine} to generate utterances for a set of words in different languages. From these utterances, the average speech embedding was computed using the Whisper large-v2 encoder. The resulting embeddings were reduced using t-SNE~\cite{van2008tsne} and plotted as shown in Figure~\ref{fig:tsne}. This initial analysis indicates that embeddings for words with the same meaning, such as \textit{``thanks''} in different languages \textit{(merci, danke, grazie, gracias)}, are closely aligned. 


\begin{figure}[!htbp]
    \centering
    \includegraphics[width=\linewidth]{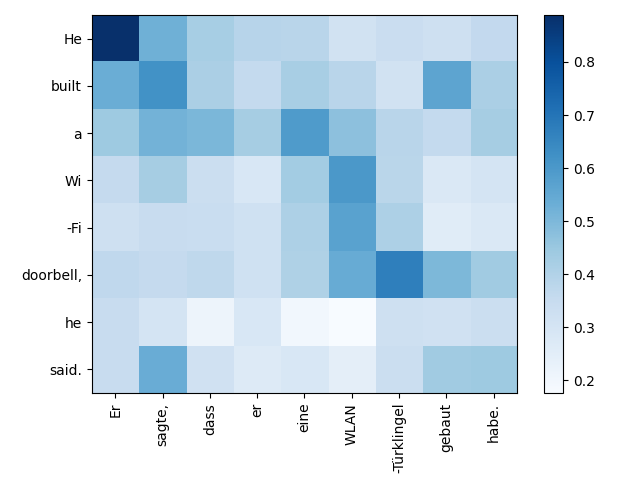}
    \caption{Cosine similarity matrix of utterance representations between an English sentence and its German counterpart selected from FLEURS test sets.
    }
    \label{fig:align}
\end{figure}

To further illustrate how languages share a common embedding space, we present an example of two parallel utterances from the FLEURS dataset, as shown in Figure \ref{fig:align}. We computed average speech embedding vectors for each word based on word-level timestamp information. The figure reveals that words with similar meanings, even if they are in different languages and have different pronunciations, tend to be mapped to similar regions in the embedding space. For instance, \textit{doorbell} (English) and \textit{Türklingel} (German) show high cosine similarity scores despite their distinct pronunciations, indicating their embeddings are close due to their shared meaning. 
Additionally, the cosine similarity matrix also reflects word order changes. For example, \textit{built} (English) and \textit{gebaut} (German) have high cosine similarity because they convey the same concept, and \textit{sagte} (German) aligns closely with \textit{said} (English). This alignment in the embedding space supports the idea that semantically similar utterances across different languages are mapped to nearby regions in the embedding space, highlighting the shared nature of the embedding space.


\subsection{Comparison of different similarity measures}
\label{app:comp_sim_measure}


To compute the similarity between two speech embedding sequences, we propose to use the AvgSim metric. The mean vector of embedding sequences $X$ and $Y$ are aggregated and then the cosine similarity between them is calculated to get an average similarity score. Compared to SeqSim, AvgSim captures the overall vector similarity rather than individual contextual speech embedding vectors. 
\begin{equation}
\small
    \text{AvgSim} = \text{CosSim}\left(\frac{1}{|X|} \sum_{\bm{x} \in X} \bm{x}, \frac{1}{|Y|} \sum_{\bm{y} \in Y} \bm{y}\right) 
\end{equation}

In Table~\ref{tab:comp}, different similarity measures are compared on three language pairs from the FLEURS data for the speech-to-speech retrieval task. Results from two additional metrics are listed here. In \cite{ctc-ot}, distance metrics based on Dynamic Time Warping (DTW) \cite{salvador2004fastdtw} and Optimal Transport (OT) \cite{coptimaltransport} are used to measure the similarity, $\text{Sim}(X, Y)$, between the contextual speech embeddings $X$ and $Y$. Both metrics use cosine distance to derive an overall sequence similarity score. 

While AvgSim is straightforward to compute, it overlooks the nuanced differences between the two sequences. DTWSim aligns the utterance representations in a monotonic fashion, which may not hold when the word order is different for the source and target sentence. To this end, we also use Optimal Transport (following \cite{ctc-ot}) to compare individual embedding pairs. We do not add a cost associated with the embedding index to ensure OT can capture token re-orderings. As the results show, it outperforms the previous two methods. Across three retrieval settings, our proposed SeqSim better captures the speech embedding similarity and shows the best performance.

\begin{table}[!htbp]
\small
    \centering
    \begin{tabular}{l|ccc}
    \toprule
        \multirow{2}*{Method} & \multicolumn{3}{c}{R@1 [\%]} \\
         & en-fr & en-de & de-fr \\
         \midrule
         Random & 0.2 & 0.2 & 0.2 \\
         \midrule
         AvgSim & 28.2 & 27.5 & 24.6 \\
         DTWSim & 29.9 & 26.5 & 22.1 \\
         OTSim & 72.3 & 66.7 & 55.2 \\
         SeqSim & \textbf{80.0} & \textbf{80.0} & \textbf{62.2} \\
    \bottomrule
    \end{tabular}
    \caption{Comparison of different similarity measures for zero-shot speech-to-speech retrieval on FLEURS.}
    \label{tab:comp}
\end{table}

\subsection{Analysis of Speech-to-Speech Retrieval}

\begin{figure}[!b]
    \centering
    \includegraphics[width=\linewidth]{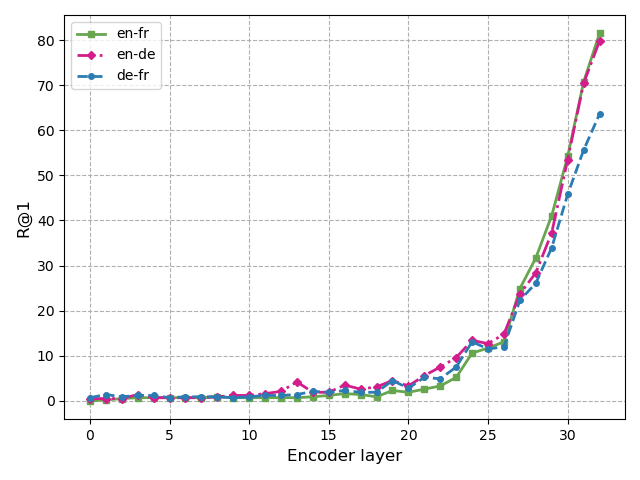}
    \caption{Speech-to-speech retrieval using outputs from different encoder layers of Whisper large-v2.}
    \label{fig:layer}
\end{figure}

In Figure \ref{fig:layer} we alternate the speech embeddings using outputs from different encoder layers of Whisper. As shown, outputs from the last encoder layer consistently achieve the best retrieval performance. For bottom layers, the recall rate drops significantly. The results indicate that outputs from higher layers are better aligned and exhibit stronger cross-lingual characteristics.

In Table~\ref{tab:model} we show the retrieval performance using encoder outputs from different Whisper models on FLEURS test sets. Even for the tiny model with only 39M parameters, the recall rate is much better than the random baseline of 0.2\%, suggesting that all models acquire the capability to do cross-lingual utterance alignment during pre-training. When the model size increases, the recall rate also improves. This implies that the retrieval performance will likely continue to improve if larger and more capable multilingual models are released in the future.
For the Whisper large models (released at different times), the v2 model shows the best performance compared to the other two versions. Whisper large-v3 is trained on additional data (5M vs 680k hours) in the form of 320k hours of weakly and 4M pseudo-labeled training data. We believe the latter degrades performance here.

\begin{table}[!t]
\small
    \centering
    \begin{tabular}{c|c|ccc}
    \toprule
        \multirow{2}*{Model} & \multirow{2}*{Size} & \multicolumn{3}{c}{R@1 [\%]} \\
         && en-fr & en-de & de-fr \\
         \midrule
         tiny & 39M & 9.2 & 9.9 & 6.8 \\
         base & 74M & 16.7 & 16.0 & 11.0 \\
         small & 244M & 27.7 & 26.1 & 20.2 \\
         medium & 769M & 50.7 & 41.8 & 39.7 \\
         \midrule
         large-v1 & \multirow{3}*{1550M} & 59.9 & 51.6 & 48.8 \\
         large-v2 && \textbf{80.0} & \textbf{80.0} & \textbf{62.2} \\
         large-v3 && 59.9 & 50.5 & 47.2 \\
    \bottomrule
    \end{tabular}
    \caption{Ablation of R@1 against different model sizes.}
    \label{tab:model}
    \vspace{-2mm}
\end{table}

In addition to FLEURS, we run speech-to-speech retrieval experiments on MaSS to validate the effectiveness of the aligned speech embedding space. Retrieval performance is presented in Table~\ref{tab:mass-more} across paired datasets in seven languages. The baseline for random selection is 0.1\% in this setting. The supervised baseline is taken from \cite{boito2020mass} who built a system based on contrastive learning \cite{harwath2018jointly}.
Excluding the low-resource language Basque (eu), the proposed zero-shot retrieval method outperforms the baseline and shows an average R@1 of 75.3\%. Although Whisper is only trained using utterances in different languages translated to English, it demonstrates good retrieval performance between arbitrary language pairs, which can be seen as an emergent ability.

\begin{table}[!htbp]
    \centering
    \small
    \begin{tabular}{@{ }c@{ }|ccccccc@{ }}
    \toprule
        \multirow{2}*{Query} & \multicolumn{7}{c}{R@1 [\%]} \\
        & en & es & ru & ro & fr & fi & eu \\
        \midrule
        en & - & 79.5 & 66.8 & 71.7 & 86.6 & 64.1 & 7.6 \\
        es & 71.9 & - & 62.7 & 83.4 & 87.5 & 62.9 & 13.4 \\
        ru & 67.8 & 72.4 & - & 83.4 & 70.4 & 72.0 & 5.5 \\
        ro & 65.5 & 84.8 & 79.1 & - & 85.1 & 69.0 & 9.7 \\
        fr & 83.0 & 91.3 & 67.0 & 89.8 & - & 66.2 & 6.9 \\
        fi & 70.1 & 74.2 & 77.4 & 81.6 & 71.7 & - & 11.2 \\
        eu & 14.6 & 25.7 & 6.5 & 14.6 & 11.3 & 9.6 & - \\
    \bottomrule
    \end{tabular}
    \caption{Zero-shot speech-to-speech retrieval results on 42 language pairs measured with SeqSim on MaSS.}
    \label{tab:mass-more}
\end{table}

\section{Ablation of Speech Translation}
\begin{table}[!t]
    \centering
    \small
    \begin{tabular}{@{ }cc|c|cc@{ }}
    \toprule
    \multirow{2}*{Dataset} & \multirow{2}*{src} & \multicolumn{3}{c}{BLEU / COMET} \\
    && FT (dec) & FT (all) & LoRA (dec) \\
    \midrule
    \multirow{4}*{FLEURS} & en & 29.1 / \textbf{78.4} & \textbf{29.3} / 77.8 & 23.3 / 73.1 \\
        & fr & \textbf{23.0} / \textbf{74.1} & 21.5 / 72.3 & 19.5 / 69.3 \\
        & de & \textbf{24.0} / \textbf{74.7} & 23.3 / 72.8 & 20.1 / 70.2 \\
        & ja & \textbf{19.2} / \textbf{74.7} & 17.7 / 72.6 & 16.8 / 72.3 \\
    \midrule
        CoVoST 2 & en & \textbf{31.9} / \textbf{76.3} & 31.2 / 75.8 & 26.3 / 72.9 \\
    \bottomrule
    \end{tabular}
    \caption{Ablation of zero-shot cross-lingual transfer.}
    \label{tab:st_ablation}
\end{table}
Ablation results are shown in Table~\ref{tab:st_ablation}. For \textit{FT~(all)}, we fine-tune all the parameters of Whisper. For \textit{LoRA~(dec)}, trainable LoRA parameters with a rank of 8 are inserted in the decoder and updated on the training set. In both settings, performance in all languages improved compared to the zero-shot results in Table~\ref{tab:st}, highlighting Whisper's effective cross-lingual transfer capability. LoRA shows worse performance compared to fine-tuning while being more parameter efficient. Moreover, compared to only fine-tuning the decoder part, fine-tuning all parameters shows similar performance on the English test set. Since the encoder parameters are changed in the adaptation, there is a shift in the speech embedding space, leading to a performance drop in languages not seen in the training. This suggests that only adapting the decoder parameters is a better strategy when extending Whisper's speech translation ability.


\begin{table}[!htbp]
    \centering
    \small
    \begin{tabular}{lc|c|c}
    \toprule
        src & code & WER & ST (zh) \\
    \midrule
        kea & pt & 89.5 & 19.5 \\
        ast & es & 47.8 & 18.7 \\
    \bottomrule
    \end{tabular}
    \caption{ASR and ST (BLEU score) into Chinese results on FLEURS data Kabuverdianu (kea) and Asturian (ast), with Whisper language code specified.
    }
    \label{tab:low_zh}
\end{table}

\begin{table*}[!htbp]
    \centering
    \small
    \begin{tabular}{c|cc|c|c|c}
    \toprule
    \multirow{2}*{src} & \multicolumn{2}{c|}{Zero-shot} & \multicolumn{3}{c}{Fine-tune}  \\
    & Translate & Transcribe & en-to-zh & fr-to-zh & ja-to-zh \\
    \midrule
    en & 1.0 / 58.8 & 10.3 / 66.3 & \textbf{19.8} / \textbf{72.2} & 18.7 / 70.6 & 14.2 / 68.1 \\
        fr & 0.9 / 56.2 & 15.7 / 66.7 & 17.0 / 68.6 & \textbf{17.1} / \textbf{68.7} & 14.4 / 66.0 \\
        de & 1.0 / 57.2 & 16.8 / 67.1 & 16.9 / \textbf{69.7} & \textbf{17.0} / 69.4 & 14.0 / 67.2 \\
        ja & 1.0 / 59.3 & 15.9 / 70.7 & 16.6 / \textbf{72.1} & 16.2 / 71.9 & \textbf{17.7} / 72.0 \\
    \bottomrule
    \end{tabular}
    \vspace{-1mm}
    \caption{Zero-shot and fine-tuning speech translation results (BLEU / COMET) for models trained on Fleurs. In the fine-tuning setup, the model is trained separately with $en\!\rightarrow\!zh$, $fr\!\rightarrow\!zh$ and $ja\!\rightarrow\!zh$ speech translation data.}
    \vspace{-2mm}
    \label{tab:fleurs_st}
\end{table*}

In Section~\ref{sec:sec:exp_st_source}, we showed that the audio embeddings for some previously unseen languages (e.g. kea and ast) align well in the shared semantic space, and these languages achieve good BLEU scores when translated into English using the baseline Whisper large-v2 model, as shown in Table~\ref{tab:low}. 
Table~\ref{tab:low_zh} demonstrates that these languages also achieve reasonable BLEU scores for Chinese translation with the fine-tuned model from Section~\ref{sec:exp_st_target} despite the high WERs.

Above, we demonstrated the expanded speech translation capabilities of Whisper by fine-tuning the model on $en\!\rightarrow\!zh$. However, one concern with this approach is the potential for catastrophic forgetting.
In Table \ref{st:en}, we study the $X\!\rightarrow\!en$ speech translation performance after the model has been fine-tuned on the $en\!\rightarrow\!zh$ training set. The results reveal a significant performance degradation when English is used as the target language, especially for languages that are more similar to Chinese. This suggests the presence of catastrophic forgetting. We aim to address this issue in our future experiments by applying elastic weight consolidation (EWC) constraints in fine-tuning \cite{kirkpatrick2017overcoming}.

\begin{table}[!htbp]
    \centering
    \small
    \begin{tabular}{c|cc}
    \toprule
        \multirow{2}*{src} & \multicolumn{2}{c}{BLEU / COMET} \\
         & before fine-tuning & after fine-tuning \\
    \midrule
         de & 37.3 / 83.4 & 22.0 / 74.7 \\
         fr & 35.1 / 83.8 & 22.8 / 75.8 \\
         ja & 18.3 / 79.2 & 5.2 / 65.2 \\
         zh & 19.7 / 80.2 & 0.1 / 65.1 \\
    \bottomrule
    \end{tabular}
    \caption{BLEU scores for Whisper models decoded on FLEURS $X\!\rightarrow\!en$ data, before and after fine-tuning on the CoVoST 2 $en\!\rightarrow\!zh$ data.}
    \label{st:en}
\end{table}

\section{Speech Translation Experiments on more $X\!\rightarrow\!Y$ directions}

Since the CoVoST 2 dataset only supports $X\!\rightarrow\!en$ and $en\!\rightarrow\!X$ translation directions, it limits our ability to experiment with more translation directions. To address this, we conducted new experiments using the FLEURS dataset, which offers n-way parallel translations. Nevertheless, it's important to note that FLEURS provides a much smaller training set compared to CoVoST 2, which may constrain the fine-tuned model's performance. In the following experiments, the target language for translation is Chinese and we used speech from three different languages as the encoder input: English, French, and Japanese. For each experiment, the training set contains 1166 utterances, contributing to around 3.5 hours of speech data. All models are trained for 20 epochs and evaluated on the same FLEURS test sets used in this paper.

Table \ref{tab:fleurs_st} presents experimental results for various speech translation setups, where speech data from different languages are utilized in the training process. As can be seen, the cross-lingual transfer learning performance depends upon the similarity between the source language used in the fine-tuning and the language of the speech to be evaluated. When Whisper is fine-tuned using English or French as the source language, similar performance gains are observed across all source languages. However, when fine-tuned with $ja\!\rightarrow\!zh$ pairs, the translation capability transfers poorly to other languages due to the substantial difference between Japanese and European languages. These findings highlight the importance of choosing a source language that closely aligns with the target language in a zero-shot transfer learning setup.

\begin{table}[!htbp]
    \centering
    \small
    \begin{tabular}{c|ccccc}
    \toprule
        src & en & fr & de & zh & ja \\
    \midrule
        BLEU & 17.3 & 5.7 & 8.5 & 4.4 & 4.8 \\
    \bottomrule
    \end{tabular}
    \caption{BLEU scores for Whisper trained on FLEURS $en\!\rightarrow\!ceb$ data and decoded on FLEURS $X\!\rightarrow\!ceb$ data.}
    \label{tab:en_ceb}
\end{table}

In Table \ref{tab:en_ceb}, we experimented with using Cebuano (ceb), a low-resource language, as the target for speech translation. Here, the training set comprises English speech with Cebuano translation annotations, containing 4.6 hours of 1262 utterances. We conducted experiments on Whisper large-v3. Since Cebuano is not supported in the Whisper speech recognition or translation pre-training, this task is more challenging compared to using Chinese as the target language. Results indicate that the model performance largely improves on the $en\!\rightarrow\!ceb$ test set after fine-tuning. Leveraging the acoustic similarity in the encoder space, translation results from other source languages show BLEU scores in a diverse range of 4.4 to 8.5. Given that the performance improvement is constrained by the limited size of the training data provided by FLEURS, we expect the model performance to improve further with the availability of a larger training set.

\end{document}